\def\ARXIV{1} 

\ifnum\ARXIV=1 
    \documentclass[sigconf]{acmart}
\else
    \documentclass[sigconf, anonymous, review]{acmart}
\fi

\usepackage{soul}
\usepackage{url}
\usepackage{amsmath}
\usepackage{amsthm}
\usepackage{booktabs}
\usepackage{algorithm}
\usepackage{algorithmic}

\usepackage{amsmath,amssymb,amsfonts}
\usepackage{algorithmic}
\usepackage{graphicx}
\usepackage{textcomp}
\usepackage{xcolor}
\usepackage{colortbl}
\usepackage{multirow}
\usepackage{booktabs}
\usepackage{makecell}

\setcopyright{none}
\renewcommand\footnotetextcopyrightpermission[1]{}

\ifnum\ARXIV=1 
    \acmConference{}{}{}
    \acmBooktitle{}
    \acmPrice{}
    \acmDOI{}
    \acmISBN{}
    \setcopyright{none}
\else
    \acmConference[CIKM '26]{Proceedings of the 35th ACM International Conference on Information and Knowledge Management}{November 7--11, 2026}{Rome, Italy}
    \acmBooktitle{Proceedings of the 35th ACM International Conference on Information and Knowledge Management (CIKM '26), November 7--11, 2026, Rome, Italy}
\fi

\ifnum\ARXIV=1 
    \title{Contrast to Detect: Dynamic Graph Contrastive Regularization for Unsupervised Anomaly Detection in Multivariate Time Series}
\else
     \title{Detecting Multivariate Time Series Anomalies through Dynamic Graph Contrastive Regularization}   
\fi

\ifnum\ARXIV=1 
    \usepackage{orcidlink}
    \author{
            Yunhua Pei$^{\dagger}$\,\orcidlink{0000-0003-2906-0827}, 
            Zixing Song$^{\ddagger}$\,\orcidlink{0000-0002-8871-3990},  
            Jin Zheng$^{\ddagger}$\,\orcidlink{0000-0002-1783-1375},  
            John Cartlidge$^{\ddagger}$\,\orcidlink{0000-0002-3143-6355}}
    \affiliation{%
      $^{\dagger}$\institution{School of Computer Science, University of Bristol \city{Bristol} \country{UK}}
      $^{\ddagger}$\institution{School of Engineering Mathematics, University of Bristol \city{Bristol}\country{UK}}
      }
    \email{
        {
        ge22472, 
        zixing.song, 
        jin.zheng, 
        john.cartlidge
        }
        @bristol.ac.uk}

\else
    \author{Yunhua Pei}
    \affiliation{%
      \institution{School of Computer Science, University of Bristol}
      \city{Bristol}
      \country{UK}}
    \email{ge22472@bristol.ac.uk}
     
    \author{Zixing Song}
    \affiliation{%
      \institution{School of Engineering Mathematics, University of Bristol}
      \city{Bristol}
      \country{UK}}
    \email{zixing.song@bristol.ac.uk}
     
    \author{Jin Zheng}
    \affiliation{%
      \institution{School of Engineering Mathematics, University of Bristol}
      \city{Bristol}
      \country{UK}}
    \email{jin.zheng@bristol.ac.uk}
     
    \author{John Cartlidge}
    \affiliation{%
      \institution{School of Engineering Mathematics, University of Bristol}
      \city{Bristol}
      \country{UK}}
    \email{john.cartlidge@bristol.ac.uk}
\fi

\ifnum\ARXIV=1
    \renewcommand{\shortauthors}{Pei, Song, Zheng, Cartlidge}
\else
    \renewcommand{\shortauthors}{Anonymous Authors}
\fi

\setcopyright{none}
\renewcommand\footnotetextcopyrightpermission[1]{}
\begin{document}

\ifnum\ARXIV=1 
    \begin{abstract}
    Anomaly detection in multivariate time series (MTS) is hindered by dynamic inter-variable dependencies and feature entanglement under spectral noise, and in practice, is further complicated by the absence of anomaly labels. Existing reconstruction-based detectors tend to recover anomalies as faithfully as normal patterns, while prevailing graph contrastive methods enforce invariance across views and thus assume a stationary relational structure, an assumption that breaks under structural drift in real systems. We propose ContrastAD, an unsupervised framework that turns structural evolution itself into a learning signal rather than suppressing it. A Multi-Perspective Embedder encodes inputs from temporal, attribute, and structural perspectives. A Frequency-Aware Attention Mixer then performs spectral top-K filtering before attention, preventing noise from leaking into query-key similarities. The core component, a Dynamic  Graph Contrastive Learner, builds power-law-inspired sparse graph snapshots from batch-level DTW distances and contrasts the most divergent pair against a stable anchor, regularizing the latent space without imposing rigid invariance.     Across five real-world benchmarks, ContrastAD attains the highest mean F1 on all five datasets and the highest AUC on three (SWaT 93.60, SMD 98.66, PSM 97.79), with statistically significant F1 and AUC margins over the strongest baseline on SWaT and PSM. On MSL and SMAP, it trails the AUC leader by under 0.7 points while still leading on F1. Ablation and sensitivity studies further confirm that the contrastive objective works best as a soft regularizer, supporting our claim that strict invariance is suboptimal under non-stationary dynamics. \textbf{Code and demo data are available online.\footnote{\url{https://anonymous.4open.science/r/CIKM26-ContrastAD-D29A}}}
    \end{abstract}
\else
    \begin{abstract}
    
    CIKM abstract here
    \\ \\ \\ \\ \\ \\ \\ \\ \\ \\
    \\ \\ \\ \\ \\ \\ \\ \\ \\ \\
    \\ \\ \\ \\ \\ \\
    \end{abstract}
\fi



\maketitle


\section{Introduction}

With the increasing complexity of modern infrastructures such as industrial control networks~\cite{yu2022edge, liu2024deep} and smart grids~\cite{zhang2021time}, reliable anomaly detection in multivariate time series (MTS) has become essential for safe and stable system operation.

The core challenges in MTS anomaly detection center on the inherent structural complexity and non-stationary nature of high-dimensional data ~\cite{deng2021graph, kim2021reversible}, which are known as (i) dynamic inter-variable dependencies and (ii) feature entanglement and spectral noise. Existing studies have made progress in different directions. Specifically, Transformer-based methods use self-attention to capture long-range temporal dependencies~\cite{zhou2021informer, zhang2025decomposition}. Frequency-domain methods attempt to untangle the overlapping periodic patterns and background noise across variables by filtering spectral noise and extracting dominant periodic components~\cite{wu2025mafcd, chen2025dual}. To better understand structural patterns, recent graph-based methods have also tried to address both challenges by representing variables as nodes and modeling their dependencies through graph structures~\cite{ding2023mst, zhao2020multivariate}. 

However, detections that are based on temporal forecasting objectives combined with topological reconstruction priors are insufficient. Firstly, these models often reconstruct anomalies as accurately as normal data~\cite{hundman2018detecting, wu2025catch}, thereby weakening their detection capability. Secondly, treating all training batches as equally reliable assumes the stationarity structure of data, even though relational patterns may drift over time as system states evolve ~\cite{lu2018learning, nam2024breaking}. Thirdly, in real-world settings, the sparsity of anomaly labels ~\cite{su2019robust} limits the practical applicability of manually supervised methods.

To address these challenges in a label-free setting, we augment reconstruction-based learning with a graph contrastive objective that enhances geometric discrimination in the latent space, separating sparse anomalies from dominant normal patterns. Moreover, real-world MTS are inherently non-stationary~\cite{nam2024breaking}, with inter-variable dependencies evolving over time, making static graph assumptions prone to false positives caused by structural drift~\cite{deng2021graph, tian2020makes}. We therefore advance to a dynamic graph contrastive framework that learns structural evolution across batches, enabling robust distinction between benign temporal shifts and genuine topological deviations. Based on this motivation, we propose ContrastAD, an unsupervised anomaly detection framework with three components: (1) a Multi-Perspective Embedder (MPE) that jointly encodes temporal, attribute-wise, and structural representations; (2) a Frequency-Aware Attention Mixer (FAM) that processes the unified embedding using temporal and frequency-aware attention for adaptive fusion; and (3) a Dynamic Graph Contrastive Learner (DGCL) that aligns naturally evolving graph snapshots. By measuring structural drift between windows, it regularizes training to avoid the rigid invariance constraints of traditional contrastive methods.

Our main contributions are summarized as follows:
\begin{itemize}
  \item We propose ContrastAD, an unsupervised anomaly detection framework that integrates multi-perspective embeddings, frequency-aware attention, and dynamic graph contrastive learning for MTS anomaly detection.
  
  \item We introduce a dynamic graph contrastive regularization strategy in ContrastAD that regularizes evolving relational patterns by leveraging divergence between consecutive graph snapshots to guide latent space optimization, rather than enforcing static invariance across views.

  \item We evaluate the framework on five public datasets against eight state-of-the-art baselines. ContrastAD achieves the highest mean F1 on all five datasets, with statistically significant F1 and AUC gains on SWaT and PSM, demonstrating robustness to sparse anomalies, class imbalance, and structural non-stationarity.
\end{itemize}

\section{Related Work}

\subsection{Multi-Perspective Anomaly Detection}
Multivariate time series (MTS) anomaly detection often requires capturing dependencies across time, variables, and latent representations. Traditional approaches typically focus on a single perspective, such as temporal modeling~\cite{zhou2021informer, zhang2025decomposition} or statistical reconstruction~\cite{hundman2018detecting}, which limits their ability to generalize under complex or evolving conditions. To address this limitation, recent studies have explored multi-view learning and multimodal feature integration. For example, DAGMM~\cite{zong2018deep} combines autoencoding with probabilistic estimation, while OmniAnomaly~\cite{su2019robust} leverages stochastic latent variables to improve generalization. More recent models, such as InterFusion~\cite{li2021multivariate}, employ hierarchical inter-metric and temporal embeddings to jointly model multi-perspective dynamics. Although these approaches improve representation capacity, challenges related to adaptive fusion and signal domain conversion noise remain, particularly when robustness is required~\cite{ma2025trishgan}.

Recent frequency-domain methods~\cite{wu2025mafcd, chen2025dual} demonstrate that many anomalies disrupt spectral patterns rather than signal magnitudes, motivating selective filtering before feature fusion~\cite{wu2025catch}. However, naively incorporating all frequency components into attention mechanisms introduces high-frequency noise that masks subtle relational deviations, underscoring the need for selective spectral filtering when integrating multiple perspectives.

\subsection{Graph-Based Anomaly Detection}

Graph-based methods have been widely adopted to model inter-variable dependencies in MTS. Early approaches construct static graphs and apply message passing using GNNs for anomaly detection~\cite{deng2021graph}. To better capture temporal variation, spatial-temporal extensions introduce multi-scale graph structures~\cite{ning2022mst}. Dynamic graph learning has also been explored to adaptively model changing correlations over time~\cite{zheng2023correlationaware, zhou2024label}, often benefiting from non-uniform or asymmetric graph structures~\cite{song2023optimal}. In addition, sparse latent graph representations have been shown to improve robustness, interpretability, and computational efficiency~\cite{han2022learning}. Despite these advances, many methods still rely on fixed or infrequently updated graphs, limiting their flexibility and effectiveness under rapidly evolving conditions. More recent detectors enhance graph attention with topological
features~\cite{liu2024multivariate} or combine dynamic graph attention with
Informer-style temporal modeling~\cite{huang2024multivariate}. Even when the
graph evolves over time, these methods use it as an input for message passing, not as successive states whose divergence drives detection.

A less explored aspect is that structural change in a graph can itself serve as an anomaly signal. When inter-variable dependencies shift due to cyber-physical attacks or workload changes, the topology reorganises in ways that static or infrequently updated models are unable to detect, which motivates approaches that explicitly track and contrast successive graph states rather than treating the graph as a fixed input.

\subsection{Contrastive Unsupervised Learning}

Contrastive learning has become a prominent self-supervised approach for learning meaningful representations without labeled data. Initial works like AD-GCL~\cite{suresh2021adversarial} and GraphCL~\cite{you2020graph} introduced graph-specific contrastive objectives leveraging data augmentations to enhance embedding quality. For time series, methods such as TS2Vec~\cite{yue2022ts2vec} and TimesURL~\cite{liu2024timesurl} extend contrastive learning to capture temporal dependencies at multiple scales. In anomaly detection, graph contrastive methods like FedCAD~\cite{kong2024federated} and GCLAD~\cite{liu2021anomaly} improve robustness and generalization by learning discriminative representations of normal pattern. However, most approaches rely on static graphs or fixed augmentations and rarely consider structural evolution as a direct learning signal, despite known degree bias and hard-negative effects in graph contrastive learning ~\cite{hu2025mitigating}, which limits their effectiveness in dynamic environments. Moreover, anomalies in MTS are typically sparse and localized, affecting only a small subset of variables or inter-variable relations~\cite{hundman2018detecting, su2019robust, deng2021graph}, which makes it difficult for models to reliably capture such subtle patterns under static or globally invariant contrastive objectives. Closely related, GraphTNC~\cite{zhang2022contrastive} contrasts temporal neighborhoods on dynamic graphs under a piecewise-stationarity assumption, and CGSTA~\cite{qi2026cgsta} aligns cross-scale graph views with a stability-aware anchor. DGCL differs by selecting the maximally divergent snapshot pair and
using its divergence as the regularization signal, without assuming local
stationarity or enforcing cross-view alignment.

A fundamental limitation of most contrastive methods is the view-invariance assumption, which trains the encoder to suppress cross-view differences and can therefore discard genuine structural change signals in non-stationary settings. Treating the contrastive objective as a soft regularizer rather than a primary learning signal offers an alternative, where a small negative weight gently separates embeddings of structurally divergent snapshots without enforcing global invariance across all views.

\begin{figure*}[!t]
    \centering
    \includegraphics[width=\textwidth]{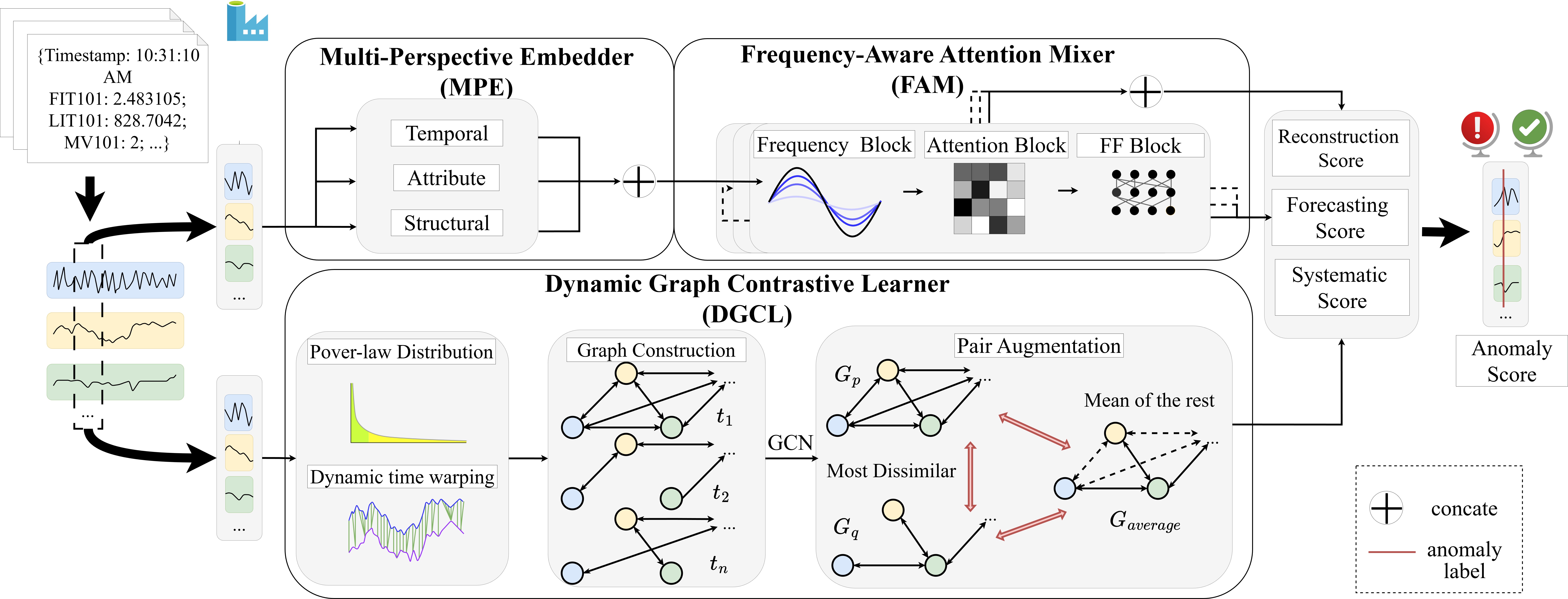}
    \caption{Overall framework of ContrastAD. The Multi-Perspective Embedder (MPE) encodes a multivariate time-series window from temporal, attribute-wise, and structural views. The Frequency-Aware Attention Mixer (FAM) applies spectral top-$K$ filtering before multi-head attention to suppress noise. In parallel, batch-level DTW distances build $S$ power-law-inspired sparse graph snapshots that feed the Dynamic Graph Contrastive Learner (DGCL). The three modules are trained jointly with forecasting and reconstruction losses and a graph-contrastive regularizer. The anomaly scores are derived from the residuals.
    }
    \Description{Architecture diagram of ContrastAD showing three modules connected in sequence: a Multi-Perspective Embedder that processes multivariate time-series segments, a Feature Attention Module that fuses temporal structural and frequency embeddings, and a Dynamic Graph Contrastive Learner that takes Dynamic Time Warping similarities to build a sparse graph. Outputs are forecasting, reconstruction, and contrastive losses combined into final anomaly scores.}
    \label{fig:framework}
\end{figure*}

\section{Methodology}
\subsection{Problem Definition}
A multivariate time series (MTS) is defined as a sequence of timestamps $\mathcal{X} = \{x_1, x_2, \dots, x_t\} \in \mathbb{R}^{N \times T}$, where $T$ is the length of the series, $N$ is the number of variables, and $x_t \in \mathbb{R}^{N}$ represents the observation vector at time $t$.

Given a training sequence assumed to be free of labels, the goal is to learn a model that captures the normal temporal dependencies and inter-signal relationships in an unsupervised manner, without using any anomaly labels. At inference time (test data), the model receives an unseen test sequence $\mathcal{X}' \in \mathbb{R}^{N \times T'}$, which may contain anomalies. The task is to predict a binary label sequence
$\widehat{\mathcal{Y}} = \{\widehat{y}_1, \widehat{y}_2, \dots, \widehat{y}_{t'}\}$ ,
where $\widehat{y}_{t'} \in \{0, 1\}$ indicates whether the observation at time $t'$
is anomalous ($1$) or normal ($0$). The predicted sequence is evaluated against the ground-truth label sequence $\mathcal{Y}$.

\subsection{Overview of ContrastAD}


The overall architecture of ContrastAD is illustrated in Fig.~\ref{fig:framework}. The model first employs a Multi-Perspective Embedder (MPE) to encode input sequences from temporal, attribute-wise, and structural perspectives, producing a unified representation that captures both temporal dynamics and inter-variable interactions. This representation is further refined by a Frequency-Aware Attention Mixer (FAM) to suppress spectral noise and enhance feature disentanglement. To handle structural drifts, the Dynamic Graph Contrastive Learner (DGCL) constructs sparse graph views from consecutive time windows and applies contrastive regularization to regularize evolving relational patterns. Finally, the learned embeddings are used for forecasting, reconstruction, and systematic consistency modeling, and anomaly scores are derived from the residuals of these tasks. A detailed block-wise illustration of FAM is provided in the supplementary material.

\subsection{Multi-Perspective Embedder (MPE)}

\paragraph{Temporal and Attribute Dilated Causal Convolutions}

Following WaveNet-style architectures~\cite{van2016wavenet, mariani2021causal, qin2022decomposed}, we adopt dilated causal convolutions (DCN) to efficiently capture long-range dependencies in MTS. The input sequence $\mathcal{X} \in \mathbb{R}^{N \times T}$ is first projected via a learnable linear transformation to $\mathbf{Z}^{(0)}$, and then processed through a stack of gated residual convolutional layers:


\begin{equation}
\begin{aligned}
    \mathbf{Z}^{(k)} = \mathbf{Z}^{(k-1)} + \mathbf{W}_{\text{res}}^{(k)} \Big( & \tanh\left(\text{Conv1D}_{r_k}^{(k)}(\mathbf{Z}^{(k-1)})\right) \\ & \odot \text{Sigmoid} \left(\text{Conv1D}_{r_k}^{(k)}(\mathbf{Z}^{(k-1)})\right) \Big), 
    \label{eq:dcn}
\end{aligned}
\end{equation}
where $k$ is the amount of DCN layers.  $r_k$ denotes the dilation rate, $\odot$ indicates element-wise multiplication, and $\mathbf{W}_{\text{res}}^{(k)}$ is the weight matrix of residual connection. 

To produce both temporal and attribute embeddings, we apply Eq.~\eqref{eq:dcn} along two orthogonal axes of the input. The temporal pass takes $\mathcal{X}$ as input and yields $z_{\text{seq}}$, encoding how each variable evolves over time. The attribute pass takes the transposed input $\mathcal{X}^{\top}$ and yields $z_{\text{fea}}$, encoding cross-variable patterns at each time step.

\paragraph{Structural Temporal Graph Attention}
To model dynamic relational dependencies across features over time, we further employ a temporal graph attention module inspired by~\cite{velivckovic2017graph}. Each feature is treated as a node in a fully connected graph, and pairwise attention scores are computed after linear projection:
\begin{equation}
    {A}_{i,j} = \text{Softmax} \left(
        \text{LeakyReLU} \left(
            {a}_{\text{src}}^\top \bar{h}_i +
            {a}_{\text{dst}}^\top \bar{h}_j
        \right)
    \right),
\end{equation}
where ${a}_{\text{src}}$ and ${a}_{\text{dst}}$ are learnable attention vectors, and $\bar{h}$ is the node feature embedding. The attention-weighted representations are then aggregated via convolution:
\begin{equation}
    {z}_{\text{gat}} = \text{Conv1D}({A} \cdot \bar{h}).
\end{equation}

Finally, representations from the three perspectives $z_{\text{seq}}$, $z_{\text{fea}}$, $z_{\text{gat}}$ are concatenated to form the unified embedding $\mathcal{Z}$. Unlike prior encoders that span at most two perspectives, such as the structural-only view in GDN~\cite{deng2021graph} or the temporal-attribute view in DTrans~\cite{qin2022decomposed}, MPE fuses all three within a single forward pass, so downstream layers can mix evidence from any of them without an extra alignment stage.

\subsection{Frequency-Aware Attention Mixer (FAM)}
Let $\mathcal{Z} \in \mathbb{R}^{L \times d}$ denote the input sequence at encoder layer $l$, 
where $L$ is the sequence length and $d$ is the embedding dimension. 
Following standard Transformer architectures~\cite{vaswani2017attention,devlin2019bert}, we first add positional encodings to obtain $\tilde{Z}$. To suppress spectral noise and emphasize dominant temporal patterns, we apply a frequency selection step prior to attention. 
Specifically, a real-valued \text{FFT}~\cite{cochran1967fast} is performed along the temporal axis, and only the top-$K$ frequency components are retained before inverse FFT reconstruction:

\begin{equation}
H_l = \text{IFFT}\left( \mathrm{Topk}(|\text{FFT}(\tilde{Z})|) \right).
\label{eq:fft_summary}
\end{equation}

Placing top-$K$ selection before the query and key projections removes spectral noise from the attention mechanism itself, rather than from its outputs after the fact. The $\mathrm{Topk}(\cdot)$ operator here acts as a hard spectral mask. In each forward pass, it zeros the real and imaginary parts of all but the $K$ highest-magnitude Fourier bins before the inverse FFT. Gradients flow through the retained bins via FFT and IFFT, then vanish on the masked bins, which is equivalent to a straight-through estimator on a binary frequency mask.

The filtered representation $H_l$ is then fed into a multi-head attention layer. For each attention head $ h $, query, key, and value matrices are computed as:
\begin{equation}
    Q_h = H_l\mathbf{W}_h^Q , \quad K_h = H_l\mathbf{W}_h^K , \quad V_h = H_l\mathbf{W}_h^V 
\end{equation}
and the scaled dot-product attention is given by:
\begin{equation}
    \text{Attention}_h = \text{Softmax}\left(\frac{Q_h K_h^\top}{\sqrt{d_k}}\right) V_h.
\end{equation}

The outputs of all heads are concatenated and projected:
\begin{equation}
    \text{MHA}(H_l) = \text{Concat}(\text{Attention}_1, \ldots, \text{Attention}_H) W^O
\end{equation}

Each encoder layer further includes a position-wise feed-forward network with residual connections and layer normalization:
\begin{equation}
    \text{FFN}(x) = \mathbf{W}_2(\text{ReLU}(\mathbf{W}_1 x + b_1))  + b_2
\end{equation}
The final embedding is obtained by a gated combination of the two FFN outputs with a residual connection.

Finally, trend and seasonal components are aggregated through linear projections and combined via a gating mechanism:
\begin{equation}
    g = \text{Sigmoid}\left( \mathbf{W}_g ( \text{FFN}_1 + \text{FFN}_2 ) \right)
\end{equation}
The final embedding is a gated sum of two processed FFN results, plus a residual connection and gating weights $\mathbf{W}_g$.

\begin{figure}[!t]
    \centering
    \includegraphics[width=0.8\columnwidth]{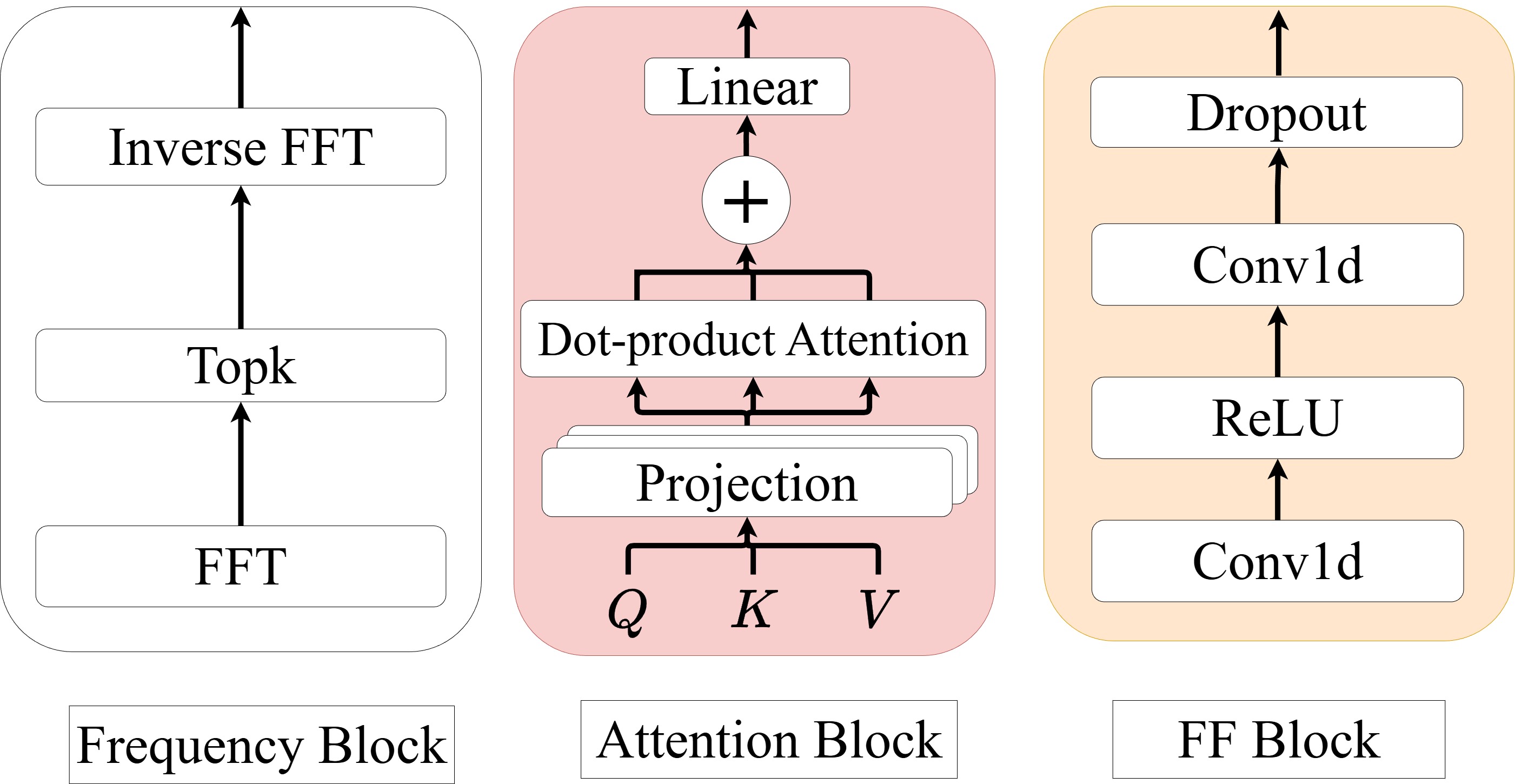}
    \caption{Structure of different blocks in FAM.}
    \label{fig:block}
\end{figure}

The overall structure of FAM is illustrated in Fig.~\ref{fig:block}: spectral filtering (Eq.~\ref{eq:fft_summary}) precedes multi-head attention, so the attention operates on a denoised, periodicity-dominant representation rather than the raw signal. This ordering is deliberate---placing FFT-based selection before attention prevents spectral noise from leaking into the query-key similarities that drive feature mixing, which we find essential for stable detection under non-stationary inputs.

\subsection{Dynamic Graph Contrastive Learner (DGCL)}

\paragraph{Sparse Graph Construction with a Power-Law Edge Budget}  
We represent the dynamic system as a sequence of graphs constructed within each sliding window $\mathbf{X}_w \in \mathbb{R}^{N \times W}$ of window size $W$, where each feature is treated as a node. To capture batch-level temporal dynamics, the time series is partitioned into \(S\) non-overlapping snapshots of equal length $\delta = W/S$.

For node \(i\) at snapshot \(s\), the corresponding time series segment is denoted as \(\mathbf{x}^i_{(s-1)\delta : s\delta} \in \mathbb{R}^\delta\). We compute pairwise Dynamic Time Warping (DTW) distances between these node segments to measure their temporal dissimilarity. The adjacency matrix \(\mathbf{A}_s \in \{0,1\}^{N \times N}\) for snapshot \(s\) is then constructed by selecting edges corresponding to the top-\(K\) DTW distances, i.e., the \(K\) most dissimilar node pairs:

\begin{equation}
A_s^{ij} =
\begin{cases}
1, & \text{if } \mathrm{DTW}\left(\mathbf{x}^i_{(s-1)\delta : s\delta},\ \mathbf{x}^j_{(s-1)\delta : s\delta}\right) \in \mathcal{T}_K \\
0, & \text{otherwise}
\end{cases}
\end{equation}
where $\mathcal{T}_K$ is the set of the $K$ largest DTW distances within snapshot $s$ (here $K$ is the per-snapshot edge budget, set in Tab.~\ref{tab:dataset_and_graph_construction}, distinct from the spectral $K$ used in FAM).

\begin{figure}[!t]
    \centering
    \includegraphics[width=0.8\columnwidth]{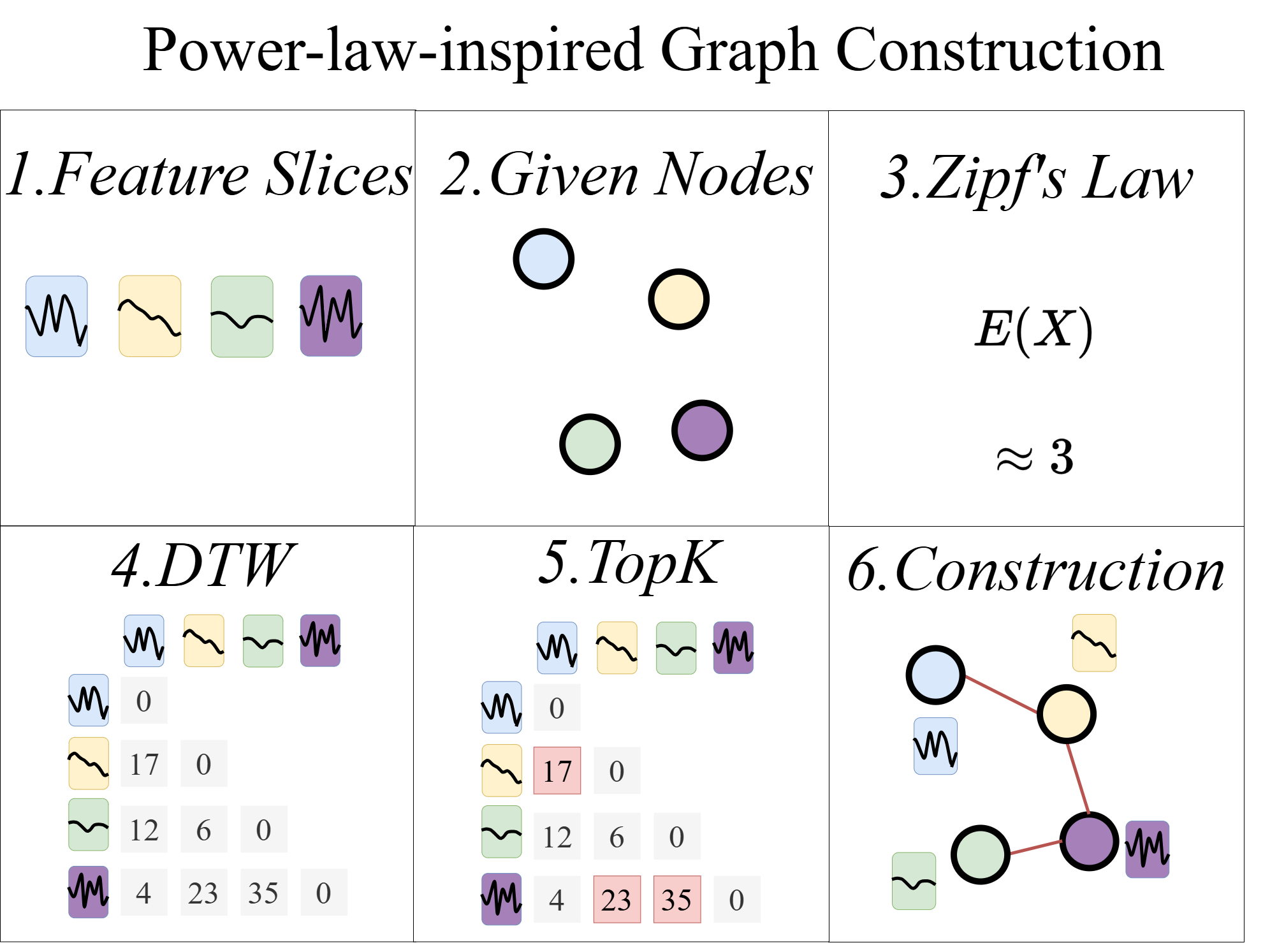}
    \caption{Example of power-law-inspired sparse graph construction in ContrastAD. Feature slices are extracted from multivariate time-series segments (Step 1), and nodes are initialized for each dimension (Step 2). A target degree distribution following Zipf’s law (Step 3) defines the expected power-law connectivity pattern $\mathcal{T}_K$. Pairwise Dynamic Time Warping (DTW) distances are computed among dimensions (Step 4), and the top-$K$ largest DTW scores are selected under the power-law edge budget (Step 5). Edges are then formed based on these selected pairs to produce a sparse adjacency graph (Step 6), which is used by DGCL.
    }
    \Description{Six-step illustration of power-law-inspired sparse graph construction. Step 1 extracts feature slices from multivariate time-series segments. Step 2 initializes one node per dimension. Step 3 shows the target Zipf-law degree distribution. Step 4 computes pairwise DTW distances between dimensions. Step 5 selects the top-K largest DTW scores under the power-law edge budget. Step 6 forms the resulting sparse adjacency graph.}
    \label{fig:graph_construction}
\end{figure}

The top-\(K\) DTW edge selection used here follows the power-law-inspired sparse graph construction strategy adopted in prior multivariate time-series work~\cite{sawhney2021exploring, icaart25}. Under this strategy, the per-snapshot edge budget follows a power-law degree prior, in which the node-degree probability decays as:
\begin{equation}
    P(n) \sim n^{-\gamma}
\end{equation}
where \(P(n)\) is the probability that a node has degree \(n\), and \(\gamma\) is the degree exponent \cite{barabasi1999emergence}. This aligns with Zipf’s law \cite{newman2005power}, where the degree probability is given by:
\begin{equation}
    P(X = n) = \frac{n^{-\gamma}}{\sum_{i=1}^N i^{-\gamma}}
\end{equation}
and \(N\) is the total number of nodes. Consequently, the expected degree of a node under this distribution is:
\begin{equation}
    E(X) = \sum_{n=1}^N n \cdot P(X = n)
    \label{eq:expected_degree}
\end{equation} 
We set the per-snapshot edge budget to $K = \lfloor N\cdot E(X)/2 \rfloor$, the expected edge count under a power-law degree prior. This prior fixes a compact, hub-aware budget but does not enforce an exact power-law degree
distribution. The edges themselves are selected by DTW dissimilarity, which keeps the graphs sparse without imposing a hard topology.

\paragraph{Contrastive Learning Objective}  
Given the set of snapshot graphs \(\mathcal{G} = \{G_1, G_2, \dots, G_S\}\), we identify the most structurally dissimilar pair \((G_p, G_q)\) by maximizing the symmetric Kullback-Leibler (KL) divergence between their normalized degree distributions:

\begin{equation}
D_{\mathrm{sym}}(\mathbf{d}_a, \mathbf{d}_b) = \sum_{k} d_a(k) \log \frac{d_a(k)}{d_b(k)} + d_b(k) \log \frac{d_b(k)}{d_a(k)},
\end{equation} 
where $k$ indexes the degree values, and $d(k)$ denotes the normalized frequency of degree $k$ in graph $G$. The most dissimilar pair is then selected as:
\begin{equation}
(p, q) = \arg\max D_{\mathrm{sym}}(\mathbf{d}_a, \mathbf{d}_b)
\end{equation}

We summarize per snapshot with a graph convolutional network (GCN) encoder.
As the DTW graph connects the most temporally divergent variables, this GCN
acts as a relational readout over a structural-contrast graph rather than a
homophily-based smoother over a similarity graph. The embeddings are computed as:



\begin{equation}
\begin{aligned}
\mathbf{z}_{a} &= \mathrm{GCN}\!\left(
\frac{1}{S - 2} \sum_{\substack{i=1 \\ i \neq p,q}}^{S} G_i
\right), \\
\mathbf{z}_{p} &= \mathrm{GCN}(G_p), \quad
\mathbf{z}_{q} = \mathrm{GCN}(G_q)
\end{aligned}
\end{equation}
where $\mathbf{z}_{p}$ and $\mathbf{z}_{q}$ are the embeddings of the two maximally divergent snapshots $G_p$ and $G_q$ (see Theoretical Motivation below).


Let $\tau>0$ denote the temperature hyperparameter. We now define an
InfoNCE-style graph contrastive regularizer. It is written in score form, and the subscripts $p$ and $q$ index the two most divergent snapshots rather than positive or negative samples. Formally, $\mathcal{L}_{\text{graph}}$ is given by:


\begin{equation}
\begin{aligned}
\label{eq:three_infonce}
\mathcal{L}_{\text{graph}}
&=
+\frac{1}{B}\sum_{i=1}^{B}
\log
\frac{
\exp(\mathbf{z}_{p,i}^{\top}\mathbf{z}_{a,i}/\tau)
}{
\sum_{j=1}^{B}\exp(\mathbf{z}_{p,i}^{\top}\mathbf{z}_{a,j}/\tau)
}
\\[4pt]
&\quad
+\frac{1}{B}\sum_{i=1}^{B}
\log
\frac{
\exp(\mathbf{z}_{q,i}^{\top}\mathbf{z}_{a,i}/\tau)
}{
\sum_{j=1}^{B}\exp(\mathbf{z}_{q,i}^{\top}\mathbf{z}_{a,j}/\tau)
}
\\[4pt]
&\quad
-\frac{1}{B}\sum_{i=1}^{B}
\log
\frac{
\exp(\mathbf{z}_{p,i}^{\top}\mathbf{z}_{q,i}/\tau)
}{
\sum_{j=1}^{B}\exp(\mathbf{z}_{p,i}^{\top}\mathbf{z}_{q,j}/\tau)
}
\end{aligned}
\end{equation}

\noindent
where the first two terms align each graph embedding with the anchor, while the third term explicitly separates embeddings of structurally dissimilar snapshots. This contrastive tension prevents representational collapse and enhances discrimination across evolving graph structures. $B$ denotes the mini-batch size of nodes.

The overall objective combines the graph contrastive regularizer with the forecasting and reconstruction losses:
\begin{equation}
\mathcal{L} = \mathcal{L}_{\text{forecast}} + \beta \cdot \mathcal{L}_{\text{reconstruction}} + \lambda \cdot \mathcal{L}_{\text{graph}}
\label{eq:total_loss}
\end{equation}
where \(\beta\) weights the reconstruction loss and \(\lambda\) the graph contrastive regularizer.

\paragraph{Theoretical Motivation of DGCL}  
We interpret DGCL through the alignment and uniformity view of contrastive
learning~\citep{wang2020understanding}. In the score form of
Eq.~\eqref{eq:three_infonce}, the first two terms grow when the two divergent snapshots stay compatible with the anchor, and the third term grows when they stay separated from each other. Maximizing the score therefore keeps each divergent snapshot compatible with the anchor while preventing the two from collapsing together, the contrastive tension DGCL relies on. Since the total objective in Eq.~\eqref{eq:total_loss} is minimized, this maximization requires a negative coefficient $\lambda$, whose small magnitude $|\lambda|\le 1$ keeps the term auxiliary to the forecasting and reconstruction losses. Standard InfoNCE objectives maximize a mutual-information lower bound~\citep{wu2020mutual, oord2018representation}, but DGCL departs from that setting, since the objective pairs two anchor-alignment terms with a snapshot-separation term and does not follow the standard InfoNCE positive and negative structure.

The power-law-inspired sparse graph construction further introduces a principled inductive bias consistent with many real-world systems, where node connectivity follows power-law distributions~\citep{barabasi1999emergence}. Moreover, graph edges are constructed using the top-$K$ highest DTW distances, connecting variables that exhibit highly dissimilar temporal behaviors. This design introduces structural diversity analogous to hard negative sampling, complementing the selection of the most dissimilar graph pair $(G_p, G_q)$ in the contrastive objective. 
Together, DTW-based hard relational edges and graph-level hard negatives enlarge the decision margin in representation space, preventing representational collapse and enhancing discrimination under temporal variation~\citep{robinson2020contrastive, kalantidis2020hard}. 
Recent theoretical analyses further suggest that optimizing InfoNCE with diverse and informative views improves downstream prediction and anomaly detection performance~\citep{saunshi2019theoretical, tian2020makes}.

\section{Experiments}
The supplementary material includes the demo code used in our experiments. Due to space constraints, we focus on reporting F1-score and AUC in the main paper, with full Precision and Recall results as well as architectural efficiency and scalability analyses provided in the supplementary material.

\subsection{Datasets}

\begin{table*}[!tbh]
\caption{Dataset statistics and graph construction.}
\label{tab:dataset_and_graph_construction}
\centering
\renewcommand{\arraystretch}{1.2}
\resizebox{0.9\textwidth}{!}{%
\begin{tabular}{l c l l r r r r r r}
\toprule
\multirow{2}{*}{\textbf{Dataset}} & \multicolumn{3}{c}{\textbf{Description}}
 & \multicolumn{4}{c}{\textbf{Statistics}} & \multicolumn{2}{c}{\textbf{Graph}} \\
\cmidrule(lr){2-4}\cmidrule(lr){5-8}\cmidrule(lr){9-10}
 & Year & Domain & Source & Train & Test & Features\,($N$) & Anomaly\,(\%) & Edges & Density\,(\%) \\
\midrule
SWaT & 2016 & Industrial control & Water treatment plant     & 475{,}200 & 449{,}919 & 51 & 12.14 & 16 & 1.25 \\
SMD  & 2019 & Server systems     & Cloud servers             & 708{,}405 & 708{,}420 & 38 & 4.16  & 13 & 1.85 \\
MSL  & 2018 & Spacecraft         & Mars rover                & 58{,}317  & 73{,}729  & 55 & 10.72 & 17 & 1.14 \\
PSM  & 2021 & Server systems     & Application servers       & 132{,}481 & 87{,}841  & 25 & 27.76 & 10 & 3.33 \\
SMAP & 2018 & Satellite          & Earth-observing satellite & 135{,}183 & 427{,}617 & 25 & 13.13 & 10 & 3.33 \\
\bottomrule
\end{tabular}%
}
\end{table*}


We evaluate our proposed framework on five real-world MTS datasets:

\begin{itemize}
    \item \textbf{SWaT}~\cite{mathur2016swat} is collected from a real-world industrial water treatment testbed that simulates a fully operational plant. It consists of 51 sensor and actuator signals sampled at one-second intervals. Anomalies correspond to cyber-attacks affecting control logic and physical processes.

    \item \textbf{SMD}~\cite{su2019robust} contains multivariate system telemetry from 28 independent server machines, each monitored for 10 days with 38 performance metrics such as CPU usage, memory consumption, and I/O throughput. Anomalies are sparse and localized, resulting in a low anomaly ratio (4.16\%).

    \item \textbf{MSL}~\cite{hundman2018detecting} is a spacecraft telemetry dataset collected from NASA’s Mars Science Laboratory rover. It contains a 55-dimensional time series reflecting sensor and actuator readings. Anomalies are rare and often manifest as subtle deviations in system dynamics rather than abrupt signal changes.

    \item \textbf{PSM}~\cite{abdulaal2021practical} originates from eBay’s application server infrastructure and includes 25 system-level performance metrics. The dataset exhibits frequent but diverse anomaly patterns caused by workload fluctuations and system faults.

    \item \textbf{SMAP}~\cite{hundman2018detecting} is a satellite telemetry dataset from NASA’s Soil Moisture Active Passive mission. It contains a 25-dimensional time series of sensor readings from an satellite. 
\end{itemize}

Tab.~\ref{tab:dataset_and_graph_construction} summarizes the detailed statistics of each dataset and the power-law-inspired constructed sparse graphs. Notably, the training data is unlabeled. For all five datasets, the training data contains no anomalies, and anomalies appear only in the test data. Training on data that also contains anomalies is left for future work. Among them, SMD exhibits a relatively sparse anomaly distribution, with only 4.16\% anomalous samples, which poses additional challenges for unsupervised anomaly detection. The undirected graph density is computed as
$\dfrac{2 \times \text{Edge}}{\text{Node} \times (\text{Node} - 1)}$, where Node denotes the number of features.







\subsection{Baselines}

We compare ContrastAD with eight representative baselines spanning graph-based and non-graph-based paradigms.

\paragraph{Graph-Based Methods}
\begin{itemize}
    \item \textbf{GDN}~\cite{deng2021graph} (AAAI 2021): A GNN-based method that dynamically constructs feature correlation graphs and detects anomalies via node-level deviations.
    \item \textbf{CSTGL}~\cite{zheng2023correlationaware} (TNNLS 2023): A correlation-aware spatial-temporal graph model that captures intra- and inter-variable dependencies using hierarchical attention.
    \item \textbf{FuSAG}~\cite{han2022learning} (KDD 2022): A graph learning model that infers sparse latent graphs across temporal and feature dimensions to capture multiscale dependencies under structured sparsity constraints.
    \item \textbf{MTG}~\cite{zhou2024label} (TKDE 2024): A label-free framework that combines multivariate temporal graphs with flow-based density estimation for anomaly detection.
    \item \textbf{MSHTrans}~\cite{chen2025mshtrans} (KDD 2025): A multi-scale hypergraph Transformer that decomposes time series into trend and seasonal components and models inter-variable relationships via learnable hyperedges.
\end{itemize}

\paragraph{Non-Graph-Based Methods}
\begin{itemize}
    \item \textbf{DTrans}~\cite{qin2022decomposed} (BigData 2022): A Transformer-based model that decomposes MTS into trend and seasonal components with frequency attention and dilated causal convolutions.
    \item \textbf{MemStr}~\cite{bhatia2022memstream} (WWW 2022): A memory-augmented streaming framework that maintains a dynamically updated bank of prototypical normal patterns without retraining.
    \item \textbf{Catch}~\cite{wu2025catch} (ICLR 2025): A channel-aware reconstruction model that patches the frequency spectrum per channel and fuses cross-channel representations via attention for anomaly detection.
\end{itemize}

\subsection{Evaluation Metrics}
Following prior work~\cite{deng2021graph, zheng2023correlationaware, han2022learning, zhou2024label}, we adopt Precision (P), Recall (R), F1-score, and AUC-ROC for evaluation, where TP, FP, and FN denote true positives, false positives, and false negatives, respectively:
\begin{equation}
P = \frac{\mathrm{TP}}{\mathrm{TP}+\mathrm{FP}}, \qquad
R = \frac{\mathrm{TP}}{\mathrm{TP}+\mathrm{FN}}, \qquad
F1 = \frac{2 \cdot P \cdot R}{P + R}.
\end{equation}
AUC-ROC measures the area under the Receiver Operating Characteristic curve, which plots the true positive rate $\mathrm{TPR}=\mathrm{TP}/(\mathrm{TP}+\mathrm{FN})$ against the false positive rate $\mathrm{FPR}=\mathrm{FP}/(\mathrm{FP}+\mathrm{TN})$ across all decision thresholds:
\begin{equation}
\mathrm{AUC} = \int_{0}^{1} \mathrm{TPR}\bigl(\mathrm{FPR}^{-1}(t)\bigr)\,\mathrm{d}t.
\end{equation}
Following standard practice in MTS anomaly detection~\cite{deng2021graph, xu2022anomaly}, we apply point adjustment uniformly to all methods before computing metrics: if any time step within a contiguous anomaly segment is correctly detected, all time steps in that segment are counted as true positives. Following recent evaluation work~\cite{kim2022towards}, the threshold-independent AUC-ROC serves as a complementary metric.

\subsection{Experimental Setup}

\begin{table*}[tbh]
\caption{Performance comparison (P: Precision, R: Recall, F1: F1-score, AUC: AUC-ROC) reported as mean\,$\pm$\,std over $n{=}5$ seeds.
Best results are in \textcolor{red}{red}, second best are \textcolor{blue}{blue}.
$^{\star}$ indicates statistically significant improvement of ContrastAD over the second-best method (non-overlapping 95\% confidence intervals, F1 and AUC only).}
\centering
\small
\resizebox{0.9\textwidth}{!}{%
\begin{tabular}{llccccccccc}
\toprule
Dataset & Metric & GDN & CSTGL & FuSAG & MTG & DTrans & MemStr & Catch & MSHTrans & ContrastAD \\
\midrule
\multirow{4}{*}{SWaT}
& P & 93.97$\pm$0.51 & 40.16$\pm$1.14 & \textcolor{red}{98.63$\pm$0.24} & 45.24$\pm$0.88 & 64.03$\pm$0.78 & 97.94$\pm$0.27 & 96.29$\pm$2.00 & \textcolor{blue}{\underline{98.15$\pm$1.83}} & 91.32$\pm$0.46 \\
& R & 71.38$\pm$1.27 & 33.12$\pm$1.33 & 71.79$\pm$1.20 & 53.50$\pm$1.09 & 71.42$\pm$0.88 & 64.30$\pm$1.18 & \textcolor{red}{74.99$\pm$2.34} & \textcolor{blue}{\underline{72.89$\pm$3.26}} & 71.72$\pm$0.50 \\
& F1 & 81.17$\pm$1.09 & 36.29$\pm$1.23 & 83.11$\pm$1.02 & 49.03$\pm$0.99 & 67.55$\pm$0.84 & 77.63$\pm$0.98 & \textcolor{blue}{\underline{84.27$\pm$1.11}} & 83.57$\pm$1.37 & \textcolor{red}{88.24$\pm$0.51$^{\star}$} \\
& AUC & 88.75$\pm$0.41 & 62.25$\pm$0.93 & 87.35$\pm$0.35 & 84.46$\pm$0.46 & 85.56$\pm$0.48 & 82.41$\pm$0.43 & \textcolor{blue}{\underline{91.94$\pm$0.17}} & 91.78$\pm$0.89 & \textcolor{red}{93.60$\pm$0.22$^{\star}$} \\
\midrule
\multirow{4}{*}{SMD}
& P & \textcolor{red}{86.39$\pm$0.27} & 17.30$\pm$0.56 & 11.91$\pm$0.38 & 18.86$\pm$0.57 & 77.17$\pm$0.50 & 11.40$\pm$0.41 & 69.58$\pm$5.90 & 78.53$\pm$1.72 & \textcolor{blue}{\underline{81.54$\pm$0.41}} \\
& R & 3.41$\pm$0.20 & 18.55$\pm$0.65 & 32.96$\pm$0.50 & 23.63$\pm$0.65 & 57.38$\pm$0.61 & 26.47$\pm$0.51 & \textcolor{blue}{\underline{78.06$\pm$5.75}} & \textcolor{red}{79.23$\pm$2.31} & 76.39$\pm$0.45 \\
& F1 & 6.55$\pm$0.33 & 17.91$\pm$0.53 & 17.50$\pm$0.46 & 20.98$\pm$0.55 & 65.87$\pm$0.57 & 15.93$\pm$0.47 & 73.57$\pm$5.82 & \textcolor{blue}{\underline{78.83$\pm$0.97}} & \textcolor{red}{78.87$\pm$0.43} \\
& AUC & 68.09$\pm$0.61 & 68.22$\pm$0.60 & 67.92$\pm$0.45 & 72.85$\pm$0.51 & 65.73$\pm$0.56 & 60.23$\pm$0.58 & 97.60$\pm$1.42 & \textcolor{blue}{\underline{98.50$\pm$0.16}} & \textcolor{red}{98.66$\pm$0.15} \\
\midrule
\multirow{4}{*}{MSL}
& P & 80.23$\pm$0.91 & 80.52$\pm$0.83 & 82.49$\pm$0.89 & \textcolor{red}{94.29$\pm$0.21} & 85.75$\pm$0.72 & 84.76$\pm$0.82 & 70.28$\pm$6.46 & 89.37$\pm$1.21 & \textcolor{blue}{\underline{90.33$\pm$0.47}} \\
& R & 75.56$\pm$0.76 & \textcolor{red}{99.44$\pm$0.14} & \textcolor{blue}{\underline{97.19$\pm$0.27}} & 16.92$\pm$0.36 & 87.43$\pm$0.61 & 95.56$\pm$0.31 & 85.25$\pm$3.73 & 86.52$\pm$1.11 & 87.50$\pm$0.43 \\
& F1 & 77.82$\pm$0.67 & 88.98$\pm$0.57 & 89.21$\pm$0.51 & 28.70$\pm$0.41 & 86.58$\pm$0.63 & \textcolor{blue}{\underline{89.84$\pm$0.55}} & 76.93$\pm$4.82 & 87.91$\pm$0.20 & \textcolor{red}{89.95$\pm$0.45} \\
& AUC & 55.77$\pm$0.82 & 57.49$\pm$0.66 & 67.23$\pm$0.62 & 61.68$\pm$0.62 & 87.87$\pm$0.53 & 83.35$\pm$0.50 & 95.36$\pm$1.10 & \textcolor{red}{98.01$\pm$0.05} & \textcolor{blue}{\underline{97.85$\pm$0.04}} \\
\midrule
\multirow{4}{*}{PSM}
& P & 59.17$\pm$0.97 & 27.77$\pm$0.48 & 52.57$\pm$0.76 & 73.47$\pm$0.61 & 76.78$\pm$0.62 & 49.41$\pm$0.65 & 80.40$\pm$15.43 & \textcolor{blue}{\underline{95.82$\pm$1.46}} & \textcolor{red}{96.87$\pm$0.40} \\
& R & 28.61$\pm$0.56 & \textcolor{red}{99.95$\pm$0.02} & 76.70$\pm$0.60 & 29.07$\pm$0.51 & 77.55$\pm$0.58 & 66.97$\pm$0.58 & 82.87$\pm$14.48 & 85.88$\pm$1.64 & \textcolor{blue}{\underline{86.21$\pm$0.35}} \\
& F1 & 38.55$\pm$0.62 & 43.46$\pm$0.78 & 62.38$\pm$0.65 & 41.64$\pm$0.58 & 77.16$\pm$0.56 & 56.86$\pm$0.62 & 78.78$\pm$0.38 & \textcolor{blue}{\underline{90.56$\pm$0.16}} & \textcolor{red}{91.17$\pm$0.11$^{\star}$} \\
& AUC & 66.90$\pm$1.04 & 55.08$\pm$0.87 & 76.41$\pm$0.42 & 72.83$\pm$0.57 & 74.86$\pm$0.55 & 69.37$\pm$0.56 & 94.22$\pm$0.25 & \textcolor{blue}{\underline{96.26$\pm$0.14}} & \textcolor{red}{97.79$\pm$0.06$^{\star}$} \\
\midrule
\multirow{4}{*}{SMAP}
& P & 8.09$\pm$0.38 & 17.30$\pm$0.60 & 17.14$\pm$0.47 & 20.58$\pm$0.48 & 14.39$\pm$0.45 & 24.94$\pm$0.53 & \textcolor{blue}{\underline{93.43$\pm$4.51}} & 88.86$\pm$2.61 & \textcolor{red}{95.63$\pm$0.51} \\
& R & 14.73$\pm$0.48 & 48.53$\pm$0.74 & \textcolor{red}{82.43$\pm$0.25} & 56.69$\pm$0.55 & 55.01$\pm$0.51 & \textcolor{blue}{\underline{64.62$\pm$0.60}} & 54.96$\pm$2.16 & 57.74$\pm$1.21 & 55.26$\pm$0.41 \\
& F1 & 10.43$\pm$0.27 & 25.50$\pm$0.55 & 28.37$\pm$0.50 & 30.20$\pm$0.46 & 22.91$\pm$0.48 & 36.00$\pm$0.51 & 69.08$\pm$0.60 & \textcolor{blue}{\underline{69.95$\pm$0.24}} & \textcolor{red}{70.04$\pm$0.02} \\
& AUC & 35.88$\pm$0.51 & 52.43$\pm$0.63 & 54.90$\pm$0.45 & 57.21$\pm$0.52 & 58.26$\pm$0.52 & 65.30$\pm$0.47 & 84.36$\pm$4.34 & \textcolor{red}{89.50$\pm$1.17} & \textcolor{blue}{\underline{88.87$\pm$0.50}} \\
\bottomrule
\end{tabular}}
\label{tab:results_combined}
\end{table*}

All experiments are conducted on an Nvidia GeForce RTX 3090 GPU with CUDA 11.5. Following prior works~\cite{deng2021graph, han2022learning, qin2022decomposed}, the data is split into training, validation, and test sets. Specifically, 80\% of the training set for training and the remaining 20\% for validation.

For MPE, we stack four DCN layers with dilation rates $r_k$ \((1, 2, 4, 8)\) to capture long-range dependencies from both temporal and attribute dimensions following ~\cite{qin2022decomposed}. For spatial modeling, we apply a temporal GAT with the same feature dimensionality, followed by a 1D convolution (kernel size 3, 32 output channels, dropout 0.1). For FAM, we stack two attention layers with 128 channels and 4 heads, using an embedding size of 128. The $K$ in Eq.~\ref{eq:fft_summary} $\mathrm{TopK}(\cdot, k)$ is 6. For DGCL, we apply a sliding window of size 100 and step size 1, the window size of 100 matches the standard choice on these five datasets~\cite{xu2022anomaly} and 10 snapshots of length 10, balancing contrastive pair diversity against DTW cost. The contrastive temperature \(\tau\) is 0.1. The forecasting loss \(\mathcal{L}_{\text{forecast}}\) and reconstruction loss \(\mathcal{L}_{\text{reconstruction}}\) are both RMSE-based. We set \(\beta = 0.1\) to balance the losses. Anomaly thresholds are selected by clustering validation scores as in~\cite{xu2022anomaly}. The contrastive loss weight $\lambda$ in $\mathcal{L}_{\text{graph}}$ is fixed per dataset prior to evaluation through pilot experiments on the training and validation splits, without consulting any test labels. The selected values are $\lambda = -0.1$ for SWaT and MSL, $\lambda = -0.9$ for SMD, $\lambda = -1.0$ for PSM, and $\lambda = -0.4$ for SMAP. Tab.~\ref{tab:sensitivity_lambda} reports a post-hoc sensitivity sweep over $\lambda \in [-1.0, +1.0]$ at intervals of 0.1 demonstrating that performance is robust around these chosen values; this sweep does not participate in $\lambda$ selection.

For each baseline, we report the hyperparameters used as follows. \textbf{GDN}~\cite{deng2021graph}: sliding window 15 with stride 5, batch size 128, embedding dimension $d=64$, top-$k=20$, one output layer with intermediate dimension 256, validation ratio 0.1, and 100 training epochs (no weight decay). \textbf{CSTGL}~\cite{zheng2023correlationaware}: input sequence length 5, batch size 4, learning rate $3 \times 10^{-4}$, weight decay $10^{-4}$, dropout 0.1, 5 training epochs, propagation $\alpha=0.1$, $\tanh\alpha=20$, 3 graph splits, node embedding dimension 256, subgraph size $k=5$, GCN depth 2, $(\text{conv}, \text{residual}, \text{skip}, \text{end})$ channels of $(16,16,32,64)$, 2 stacked layers, with 5 PCA components and a normalization window of 200. \textbf{FuSAG}~\cite{han2022learning}: sliding window 5 with stride 1, batch size 32, embedding dimension $d=64$, top-$k=15$, output intermediate dimension 64, validation ratio 0.2, learning rate $10^{-3}$, forecasting loss weight $\alpha=0.5$, and sparse loss weight $\beta=1.0$. \textbf{MTG}~\cite{zhou2024label}: window size 60 with stride 10, batch size 128, learning rate $2 \times 10^{-3}$, weight decay $5 \times 10^{-4}$, MAF architecture with 1 block, 1 hidden layer of size 32, and 1 Gaussian component. \textbf{MSHTrans}~\cite{chen2025mshtrans}: temporal scales $(1, 2, 4)$, model dimension $d=64$, 16 learnable hyperedges, 4 attention heads, and dropout 0.1. \textbf{DTrans}~\cite{qin2022decomposed}: window size 100, batch size 128, learning rate $10^{-4}$, dropout 0.2, kernel size 3, 1 GRU layer, 1 forecast layer with hidden dimension 150, and reconstruction/anomaly weights $\lambda = \beta = 1$. \textbf{MemStr}~\cite{bhatia2022memstream}: memory length 2048, autoencoder pre-training for 5000 epochs with learning rate $10^{-2}$, and threshold $\beta=0.1$. \textbf{Catch}~\cite{wu2025catch}: patch size 10, model dimension $d=64$, 4 attention heads, 2 encoder layers, and dropout 0.1. All baselines are evaluated under identical point-adjustment settings and the five-seed protocol as ContrastAD $\{0, 1, 2, 3, 42\}$.

\subsection{Results and Analysis}

\begin{table*}[!tbh]
\caption{Ablation study results are reported as mean\,$\pm$\,std over $n{=}5$ seeds. We evaluate the full model (ContrastAD) and its variants by ablating the MPE (\textbackslash M), the FAM (\textbackslash F), and the DGCL (\textbackslash D). Best results are in \textcolor{red}{red}, second best are \textcolor{blue}{blue}. $^{\star}$ indicates statistically significant improvement over the best variant (non-overlapping 95\% confidence intervals).}
\centering
\small
\resizebox{\textwidth}{!}{%
\begin{tabular}{lcccccccccc}
\toprule
\multirow{2}{*}{Variant}
& \multicolumn{2}{c}{SWaT}
& \multicolumn{2}{c}{SMD}
& \multicolumn{2}{c}{MSL}
& \multicolumn{2}{c}{PSM}
& \multicolumn{2}{c}{SMAP} \\
\cmidrule(lr){2-3} \cmidrule(lr){4-5} \cmidrule(lr){6-7}
\cmidrule(lr){8-9} \cmidrule(lr){10-11}
& F1 & AUC & F1 & AUC & F1 & AUC & F1 & AUC & F1 & AUC \\
\midrule
ContrastAD\textbackslash M
& \textcolor{blue}{80.77$\pm$0.70} & \textcolor{blue}{90.22$\pm$0.38}
& 38.01$\pm$0.51 & 62.73$\pm$0.46
& 79.46$\pm$0.62 & \textcolor{blue}{77.81$\pm$0.41}
& 73.85$\pm$0.58 & 73.92$\pm$0.53
& 28.17$\pm$0.46 & 59.13$\pm$0.48 \\

ContrastAD\textbackslash F
& 77.65$\pm$0.68 & 88.81$\pm$0.36
& 47.11$\pm$0.48 & \textcolor{blue}{72.60$\pm$0.43}
& \textcolor{blue}{89.03$\pm$0.55} & 76.15$\pm$0.37
& \textcolor{blue}{81.17$\pm$0.57} & \textcolor{blue}{74.93$\pm$0.52}
& 30.79$\pm$0.41 & \textcolor{blue}{61.97$\pm$0.45} \\

ContrastAD\textbackslash D
& 73.84$\pm$0.73 & 81.90$\pm$0.42
& \textcolor{blue}{50.01$\pm$0.47} & 69.51$\pm$0.45
& 84.65$\pm$0.58 & 69.95$\pm$0.40
& 80.84$\pm$0.52 & 70.33$\pm$0.48
& \textcolor{blue}{46.43$\pm$0.43} & 60.69$\pm$0.46 \\

\midrule
ContrastAD
& \textcolor{red}{88.24$\pm$0.51}$^{\star}$ & \textcolor{red}{93.60$\pm$0.22}$^{\star}$
& \textcolor{red}{78.87$\pm$0.43}$^{\star}$ & \textcolor{red}{98.66$\pm$0.15}$^{\star}$
& \textcolor{red}{89.95$\pm$0.45} & \textcolor{red}{97.85$\pm$0.04}$^{\star}$
& \textcolor{red}{91.17$\pm$0.11}$^{\star}$ & \textcolor{red}{97.79$\pm$0.06}$^{\star}$
& \textcolor{red}{70.04$\pm$0.02}$^{\star}$ & \textcolor{red}{88.87$\pm$0.50}$^{\star}$ \\
\bottomrule
\end{tabular}}
\label{tab:ablation_all_ci}
\end{table*}

\begin{table*}[tbh]
\caption{Sensitivity of the contrastive weight $\lambda$ on F1 and AUC-ROC (mean$\pm$std, $n\!=\!5$, values $\!\times\!100$). Each column is independently shaded from \colorbox[HTML]{FFFFFF}{\strut worst} to {\setlength{\fboxsep}{1pt}\colorbox[HTML]{7F2704}{\textcolor{white}{\strut best}}}. The \colorbox[HTML]{E8E8E8}{\strut shaded} row ($\lambda\!=\!0$) removes DGCL.}
\label{tab:sensitivity_lambda}
\centering\scriptsize
\setlength{\tabcolsep}{3pt}
\renewcommand{\arraystretch}{1.0}
\resizebox{0.9\textwidth}{!}{%
\begin{tabular}{rcccccccccc}
\toprule
\textbf{$\lambda$} & \multicolumn{2}{c}{\textbf{SWaT}} & \multicolumn{2}{c}{\textbf{SMD}} & \multicolumn{2}{c}{\textbf{MSL}} & \multicolumn{2}{c}{\textbf{PSM}} & \multicolumn{2}{c}{\textbf{SMAP}} \\
\cmidrule(lr){1-1}\cmidrule(lr){2-3}\cmidrule(lr){4-5}\cmidrule(lr){6-7}\cmidrule(lr){8-9}\cmidrule(lr){10-11}
 & F1 & AUC & F1 & AUC & F1 & AUC & F1 & AUC & F1 & AUC \\
\midrule
$-1.0$ & \cellcolor[HTML]{EF7933}82.64{$\pm$1.47} & 90.68{$\pm$0.97} & \cellcolor[HTML]{FDCC9F}78.01{$\pm$0.68} & 98.55{$\pm$0.14} & \cellcolor[HTML]{E0520C}\textcolor{white}{88.81{$\pm$0.25}} & 98.15{$\pm$0.07} & \cellcolor[HTML]{7F2704}\textcolor{white}{91.17{$\pm$0.11}} & 97.79{$\pm$0.06} & \cellcolor[HTML]{FDBF89}70.01{$\pm$0.06} & 89.43{$\pm$1.02} \\
$-0.9$ & \cellcolor[HTML]{FEFBF7}75.42{$\pm$12.77} & 86.46{$\pm$6.02} & \cellcolor[HTML]{7F2704}\textcolor{white}{78.87{$\pm$1.12}} & 98.66{$\pm$0.15} & \cellcolor[HTML]{FDEFDE}88.31{$\pm$0.79} & 98.08{$\pm$0.12} & \cellcolor[HTML]{FDCC9F}90.20{$\pm$1.23} & 97.72{$\pm$0.30} & \cellcolor[HTML]{E0520C}\textcolor{white}{70.03{$\pm$0.07}} & 89.21{$\pm$1.04} \\
$-0.8$ & \cellcolor[HTML]{F9A25E}82.28{$\pm$0.80} & 88.50{$\pm$1.53} & \cellcolor[HTML]{FDEAD5}77.73{$\pm$0.60} & 98.55{$\pm$0.15} & \cellcolor[HTML]{FEF7EF}88.21{$\pm$0.32} & 98.08{$\pm$0.06} & \cellcolor[HTML]{EA651E}90.60{$\pm$0.99} & 97.72{$\pm$0.20} & \cellcolor[HTML]{E0520C}\textcolor{white}{70.03{$\pm$0.07}} & 89.27{$\pm$0.66} \\
$-0.7$ & \cellcolor[HTML]{FDCC9F}81.63{$\pm$3.81} & 89.23{$\pm$2.47} & \cellcolor[HTML]{AE3C08}\textcolor{white}{78.47{$\pm$0.78}} & 98.59{$\pm$0.15} & \cellcolor[HTML]{FDCC9F}88.58{$\pm$0.19} & 98.10{$\pm$0.07} & \cellcolor[HTML]{FDE4CA}90.04{$\pm$1.13} & 97.67{$\pm$0.12} & \cellcolor[HTML]{FDD8B4}70.00{$\pm$0.05} & 89.03{$\pm$0.52} \\
$-0.6$ & \cellcolor[HTML]{FDE4CA}81.05{$\pm$0.95} & 88.68{$\pm$1.93} & \cellcolor[HTML]{FDEFDE}77.71{$\pm$0.32} & 98.56{$\pm$0.11} & \cellcolor[HTML]{FDBF89}88.62{$\pm$0.13} & 98.12{$\pm$0.02} & \cellcolor[HTML]{FDEAD5}90.03{$\pm$0.32} & 97.65{$\pm$0.18} & \cellcolor[HTML]{FDBF89}70.01{$\pm$0.02} & 89.11{$\pm$0.65} \\
$-0.5$ & \cellcolor[HTML]{FDD8B4}81.31{$\pm$0.75} & 89.68{$\pm$1.39} & \cellcolor[HTML]{FDB374}78.09{$\pm$0.37} & 98.55{$\pm$0.10} & \cellcolor[HTML]{FDE4CA}88.47{$\pm$0.32} & 98.11{$\pm$0.09} & \cellcolor[HTML]{FDD8B4}90.08{$\pm$1.08} & 97.75{$\pm$0.15} & \cellcolor[HTML]{F9A25E}70.02{$\pm$0.04} & 88.77{$\pm$0.51} \\
$-0.4$ & \cellcolor[HTML]{E0520C}\textcolor{white}{83.26{$\pm$1.64}} & 90.02{$\pm$1.63} & \cellcolor[HTML]{FDE4CA}77.76{$\pm$0.59} & 98.58{$\pm$0.10} & \cellcolor[HTML]{EF7933}88.73{$\pm$0.71} & 98.16{$\pm$0.06} & \cellcolor[HTML]{D54D0B}\textcolor{white}{90.69{$\pm$1.03}} & 97.75{$\pm$0.24} & \cellcolor[HTML]{7F2704}\textcolor{white}{70.04{$\pm$0.02}} & 88.87{$\pm$0.50} \\
$-0.3$ & \cellcolor[HTML]{FDEFDE}79.22{$\pm$7.08} & 88.92{$\pm$3.46} & \cellcolor[HTML]{F9A25E}78.12{$\pm$0.86} & 98.53{$\pm$0.12} & \cellcolor[HTML]{FDEAD5}88.41{$\pm$0.37} & 98.11{$\pm$0.12} & \cellcolor[HTML]{FDBF89}90.39{$\pm$0.63} & 97.82{$\pm$0.12} & \cellcolor[HTML]{E0520C}\textcolor{white}{70.03{$\pm$0.03}} & 88.78{$\pm$0.40} \\
$-0.2$ & \cellcolor[HTML]{AE3C08}\textcolor{white}{84.60{$\pm$12.38}} & 87.62{$\pm$5.70} & \cellcolor[HTML]{FEF7EF}74.06{$\pm$0.40} & 98.33{$\pm$0.14} & \cellcolor[HTML]{963106}\textcolor{white}{89.13{$\pm$0.44}} & 97.70{$\pm$0.08} & \cellcolor[HTML]{AE3C08}\textcolor{white}{90.95{$\pm$0.93}} & 97.79{$\pm$0.28} & \cellcolor[HTML]{FEF7EF}48.05{$\pm$0.03} & 88.24{$\pm$0.61} \\
$-0.1$ & \cellcolor[HTML]{7F2704}\textcolor{white}{88.24{$\pm$5.00}} & 93.60{$\pm$0.22} & \cellcolor[HTML]{FDF2E6}76.81{$\pm$0.45} & 98.36{$\pm$0.15} & \cellcolor[HTML]{7F2704}\textcolor{white}{89.95{$\pm$0.18}} & 97.85{$\pm$0.04} & \cellcolor[HTML]{FCAC69}90.41{$\pm$1.35} & 97.68{$\pm$0.31} & \cellcolor[HTML]{FDF2E6}49.08{$\pm$0.05} & 87.85{$\pm$0.57} \\
\midrule
\textbf{$0$} & \cellcolor[HTML]{E8E8E8}73.84{$\pm$0.73} & \cellcolor[HTML]{E8E8E8}81.90{$\pm$0.42} & \cellcolor[HTML]{E8E8E8}50.01{$\pm$0.47} & \cellcolor[HTML]{E8E8E8}69.51{$\pm$0.45} & \cellcolor[HTML]{E8E8E8}84.65{$\pm$0.58} & \cellcolor[HTML]{E8E8E8}69.95{$\pm$0.40} & \cellcolor[HTML]{E8E8E8}80.84{$\pm$0.52} & \cellcolor[HTML]{E8E8E8}70.33{$\pm$0.48} & \cellcolor[HTML]{E8E8E8}46.43{$\pm$0.43} & \cellcolor[HTML]{E8E8E8}60.69{$\pm$0.46} \\
\midrule
$+0.1$ & \cellcolor[HTML]{963106}\textcolor{white}{85.53{$\pm$2.21}} & 89.39{$\pm$1.34} & \cellcolor[HTML]{FEFBF7}72.63{$\pm$0.44} & 98.40{$\pm$0.10} & \cellcolor[HTML]{FEFBF7}86.65{$\pm$0.38} & 98.16{$\pm$0.11} & \cellcolor[HTML]{FEFBF7}88.45{$\pm$1.09} & 97.76{$\pm$0.17} & \cellcolor[HTML]{FEFBF7}47.34{$\pm$0.06} & 88.20{$\pm$0.45} \\
$+0.2$ & \cellcolor[HTML]{FEF7EF}76.70{$\pm$4.69} & 88.68{$\pm$2.00} & \cellcolor[HTML]{FFFFFF}68.88{$\pm$0.48} & 98.70{$\pm$0.14} & \cellcolor[HTML]{FFFFFF}80.58{$\pm$0.54} & 98.06{$\pm$0.13} & \cellcolor[HTML]{FFFFFF}86.13{$\pm$0.94} & 97.87{$\pm$0.16} & \cellcolor[HTML]{FFFFFF}39.39{$\pm$0.06} & 88.22{$\pm$0.48} \\
$+0.3$ & \cellcolor[HTML]{FDF2E6}77.33{$\pm$8.69} & 87.22{$\pm$3.23} & \cellcolor[HTML]{963106}\textcolor{white}{78.52{$\pm$0.52}} & 98.62{$\pm$0.11} & \cellcolor[HTML]{F9A25E}88.71{$\pm$0.12} & 98.17{$\pm$0.05} & \cellcolor[HTML]{963106}\textcolor{white}{91.14{$\pm$1.03}} & 97.81{$\pm$0.19} & \cellcolor[HTML]{FDBF89}70.01{$\pm$0.05} & 88.22{$\pm$0.65} \\
$+0.4$ & \cellcolor[HTML]{FFFFFF}73.90{$\pm$16.41} & 85.95{$\pm$8.27} & \cellcolor[HTML]{E0520C}\textcolor{white}{78.34{$\pm$0.70}} & 98.58{$\pm$0.14} & \cellcolor[HTML]{FDF2E6}88.22{$\pm$0.91} & 98.05{$\pm$0.15} & \cellcolor[HTML]{FEF7EF}89.69{$\pm$1.16} & 97.58{$\pm$0.48} & \cellcolor[HTML]{E0520C}\textcolor{white}{70.03{$\pm$0.03}} & 87.76{$\pm$0.49} \\
$+0.5$ & \cellcolor[HTML]{F48E49}82.55{$\pm$2.28} & 89.49{$\pm$1.34} & \cellcolor[HTML]{FDD8B4}77.94{$\pm$0.61} & 98.62{$\pm$0.09} & \cellcolor[HTML]{F48E49}88.72{$\pm$0.12} & 98.10{$\pm$0.05} & \cellcolor[HTML]{FDF2E6}89.72{$\pm$1.23} & 97.64{$\pm$0.29} & \cellcolor[HTML]{E0520C}\textcolor{white}{70.03{$\pm$0.04}} & 87.78{$\pm$0.57} \\
$+0.6$ & \cellcolor[HTML]{C8470A}\textcolor{white}{83.39{$\pm$1.44}} & 90.21{$\pm$0.67} & \cellcolor[HTML]{ED7029}78.30{$\pm$0.45} & 98.60{$\pm$0.10} & \cellcolor[HTML]{AE3C08}\textcolor{white}{88.86{$\pm$0.17}} & 98.16{$\pm$0.07} & \cellcolor[HTML]{F48E49}90.43{$\pm$1.33} & 97.73{$\pm$0.14} & \cellcolor[HTML]{FDEAD5}65.47{$\pm$0.06} & 87.51{$\pm$0.78} \\
$+0.7$ & \cellcolor[HTML]{FDB374}82.23{$\pm$1.55} & 89.11{$\pm$1.76} & \cellcolor[HTML]{FDBF89}78.06{$\pm$0.31} & 98.58{$\pm$0.13} & \cellcolor[HTML]{EA651E}88.78{$\pm$0.79} & 98.09{$\pm$0.09} & \cellcolor[HTML]{FCAC69}90.41{$\pm$0.33} & 97.72{$\pm$0.12} & \cellcolor[HTML]{E0520C}\textcolor{white}{70.03{$\pm$0.05}} & 87.67{$\pm$0.62} \\
$+0.8$ & \cellcolor[HTML]{EA651E}82.89{$\pm$1.90} & 89.33{$\pm$0.67} & \cellcolor[HTML]{ED7029}78.30{$\pm$0.45} & 98.56{$\pm$0.10} & \cellcolor[HTML]{FDB374}88.65{$\pm$0.22} & 98.10{$\pm$0.07} & \cellcolor[HTML]{FDEFDE}89.77{$\pm$0.37} & 97.59{$\pm$0.22} & \cellcolor[HTML]{FDEFDE}63.22{$\pm$0.04} & 87.62{$\pm$0.78} \\
$+0.9$ & \cellcolor[HTML]{FDEAD5}80.61{$\pm$4.63} & 88.83{$\pm$2.86} & \cellcolor[HTML]{F48E49}78.25{$\pm$0.52} & 98.60{$\pm$0.11} & \cellcolor[HTML]{FDD8B4}88.56{$\pm$0.28} & 98.11{$\pm$0.07} & \cellcolor[HTML]{EF7933}90.45{$\pm$1.13} & 97.74{$\pm$0.16} & \cellcolor[HTML]{E0520C}\textcolor{white}{70.03{$\pm$0.03}} & 87.44{$\pm$1.38} \\
$+1.0$ & \cellcolor[HTML]{FDBF89}81.71{$\pm$1.47} & 88.69{$\pm$1.89} & \cellcolor[HTML]{C8470A}\textcolor{white}{78.40{$\pm$0.34}} & 98.64{$\pm$0.09} & \cellcolor[HTML]{C8470A}\textcolor{white}{88.85{$\pm$0.19}} & 98.10{$\pm$0.11} & \cellcolor[HTML]{D54D0B}\textcolor{white}{90.69{$\pm$1.11}} & 97.79{$\pm$0.18} & \cellcolor[HTML]{FDE4CA}68.18{$\pm$0.10} & 87.26{$\pm$1.34} \\
\bottomrule
\end{tabular}}
\end{table*}


\subsubsection{RQ1. Performance Comparison} 
Tab.~\ref{tab:results_combined} summarizes results across eight baselines and five datasets. ContrastAD attains the highest mean F1 on all five datasets and the highest AUC on three of them(SWaT, SMD, and PSM), with statistically significant F1 and AUC margins on SWaT and PSM, while MSHTrans leads on AUC for MSL and SMAP. The pattern of F1 leadership across every dataset paired with AUC leadership on three out of five supports DGCL's role as a boundary-shaping regularizer that improves both operating-threshold quality and global score separation.

On SWaT, ContrastAD reaches F1 88.24\% ($^{\star}$, +3.97 over Catch) and AUC 93.60\% ($^{\star}$, +1.66 over Catch). The wider F1 margin relative to the AUC margin indicates that DGCL's main contribution on coordinated multi-variable attacks is sharpening the operating threshold rather than uniformly widening score separation. On SMD (4.16\% anomaly ratio), ContrastAD edges out MSHTrans on both F1 (+0.04) and AUC (+0.16), showing that DGCL's relational regularizer remains effective on sparse anomalies even when the margin to the strongest baseline is tight. On PSM, ContrastAD achieves both the best F1 ($^{\star}$, 91.17\%) and AUC ($^{\star}$, 97.79\%), suggesting that diverse anomaly patterns benefit especially from relational evolution modelling. Several baselines collapse to degenerate precision-recall tradeoffs. GDN on SMD yields F1 of only 6.55\% despite 86.39\% Precision (Recall 3.41\%), and CSTGL on PSM yields 43.46\% F1 despite 99.95\% Recall (Precision 27.77\%). ContrastAD avoids both failure modes on every dataset, which is the direct result of DGCL tightening the normal manifold without over-committing to either detection direction.

\subsubsection{RQ2. Ablation Study}

\begin{table*}[tbh]
\caption{Architectural and computational characteristics of different anomaly detection methods on the SWaT dataset. $N_L$ denotes test-sequence length. Performance tiers use \checkmark~(low), \checkmark\checkmark~(medium), and \checkmark\checkmark\checkmark~(high) based on aggregate average F1 and AUC across all datasets. Complexities are reported under the fixed sliding window ($L=100$).}
\label{tab:method_characteristics}
\centering
\small
\resizebox{0.9\textwidth}{!}{%
\begin{tabular}{llccccc}
\toprule
\textbf{Type} &
\textbf{Method} &
\textbf{Train Parallel} &
\textbf{Inference Cost} &
\textbf{Memory Complexity} &
\textbf{Train Time (s/epoch)} &
\textbf{Perf. Tier} \\
\midrule
\multirow{5}{*}{\textit{Graph-based}} &
GDN~\cite{deng2021graph} & \checkmark & $\mathcal{O}(1)$ & $\mathcal{O}(N_L)$ & 9.53 & \checkmark\checkmark \\
& CSTGL~\cite{zheng2023correlationaware} & \checkmark & $\mathcal{O}(1)$ & $\mathcal{O}(N_L)$ & 384.53 & \checkmark \\
& FuSAG~\cite{han2022learning} & \checkmark & $\mathcal{O}(1)$ & $\mathcal{O}(N_L)$ & 156.22 & \checkmark\checkmark \\
& MTG~\cite{zhou2024label} & \checkmark & $\mathcal{O}(1)$ & $\mathcal{O}(N_L)$ & 2.80 & \checkmark \\
& MSHTrans~\cite{chen2025mshtrans} & \checkmark & $\mathcal{O}(1)$ & $\mathcal{O}(N_L^2)$ & 634.72 & \checkmark\checkmark\checkmark \\
\midrule
\multirow{3}{*}{\textit{Non-graph-based}} &
DTrans~\cite{qin2022decomposed} & \checkmark & $\mathcal{O}(N_L)$ & $\mathcal{O}(N_L^2)$ & 141.64 & \checkmark\checkmark \\
& MemStr~\cite{bhatia2022memstream} & -- & $\mathcal{O}(1)$ & $\mathcal{O}(N_L)$ & 96.39 & \checkmark\checkmark \\
& Catch~\cite{wu2025catch} & \checkmark & $\mathcal{O}(1)$ & $\mathcal{O}(N_L^2)$ & 190.49 & \checkmark\checkmark\checkmark \\
\midrule
\textit{Ours} & ContrastAD & \checkmark & $\mathcal{O}(1)$ & $\mathcal{O}(N_L)$ & 132.16 & \checkmark\checkmark\checkmark \\
\bottomrule
\end{tabular}}
\end{table*}

To assess the contribution of each core component, we replace MPE (\textbackslash M) with a linear projection, FAM (\textbackslash F) with standard multi-head attention, and exclude DGCL (\textbackslash D) entirely. Results are in Tab.~\ref{tab:ablation_all_ci}.

Every ablation degrades performance consistently across datasets. Removing DGCL (\textbackslash D) causes the most striking single-module drop on SWaT, where F1 falls from 88.24 to 73.84, equal to the $\lambda\!=\!0$ sensitivity baseline, confirming that DGCL's contribution cannot be recovered by the reconstruction objective alone. Removing MPE (\textbackslash M) produces the largest absolute drop on SMD (from 78.87 to 38.01), suggesting that multi-perspective representation is especially critical when anomalies are sparse and non-stationary structural shifts provide the primary detection signal. Replacing FAM (\textbackslash F) with standard attention degrades performance on SMAP and SMD most noticeably, indicating that frequency-aware filtering before attention is not interchangeable with generic mixing. Taken together, the ablation reveals that the three modules address complementary failure modes rather than redundant functions, which explains why their gains are additive rather than overlapping.

\subsubsection{RQ3. Hyperparameter Sensitivity}



Tab.~\ref{tab:sensitivity_lambda} reports F1 and AUC across $\lambda \in [-1.0, +1.0]$ at 0.1 intervals. Setting $\lambda\!=\!0$ removes DGCL entirely and uniformly produces the lowest F1 on every dataset, with drops as large as 28.86 points on SMD, establishing that the contrastive signal is indispensable and cannot be absorbed by the reconstruction loss. Any non-zero $\lambda$ substantially restores performance, and negative values outperform their positive counterparts on average across all five datasets. For SWaT and MSL, performance peaks at $\lambda\!=\!-0.1$, while SMD and PSM favour larger magnitudes ($-0.9$ and $-1.0$ respectively), reflecting how the degree of structural non-stationarity varies across domains. The key insight from this table is that performance remains largely stable across the full non-zero range, with F1 standard deviations well below 2 points on most datasets outside the $\lambda\!=\!0$ singularity. This robustness confirms that DGCL acts as a soft regularizer whose benefit does not depend on precise tuning, and that the critical design choice is the sign of $\lambda$ rather than its magnitude.

The sign of $\lambda$ follows directly from this score form and the
alignment-uniformity view of contrastive learning~\citep{wang2020understanding}. A negative $\lambda$ maximizes the graph contrastive regularizer and produces the contrastive tension above, whereas a positive $\lambda$ minimizes it, collapses the divergent snapshots, and recovers the static invariance the model is designed to avoid. Tab.~\ref{tab:sensitivity_lambda} matches this picture on three fronts. AUC at $\lambda\!=\!0$ stays below 71 on SMD, MSL, PSM, and SMAP and recovers above 87 once any non-zero $\lambda$ is added, the best F1 on every dataset sits at a negative $\lambda$, and the datasets with the strongest structural non-stationarity prefer the largest magnitudes (SMD F1 rises 28.86 points and AUC rises from 69.51 to 98.66 between $\lambda\!=\!0$ and $-0.9$). The take-away is that the sign flip restores separability, and magnitude only tunes how aggressively diversity is preserved.

\subsubsection{RQ4. Architectural Efficiency and Scalability Analysis}

Tab.~\ref{tab:method_characteristics} compares training parallelism, inference cost, memory complexity, and per-epoch training time. Most graph-based baselines support parallel training with linear memory but either incur high cost (CSTGL at 384.53s, FuSAG at 156.22s, MSHTrans at 634.72s) or fall short on detection (MTG). DTrans achieves competitive accuracy but requires $\mathcal{O}(N_L^2)$ memory due to global attention, and MemStr sacrifices parallelism for streaming efficiency. Catch and MSHTrans reach the high performance tier but both require $\mathcal{O}(N_L^2)$ memory due to self-attention, which limits their applicability as the window length $N_L$ grows. ContrastAD is the only method in the high performance tier that simultaneously supports parallel training, $\mathcal{O}(1)$ inference, and $\mathcal{O}(N_L)$ memory. This favourable profile arises from a design choice that may not be immediately obvious. DGCL operates on sparse graph snapshots with a fixed number of nodes, so its memory footprint scales with the variable count $N$ rather than the window length $N_L$, unlike attention-based approaches that must materialise an $N_L \times N_L$ similarity matrix at every layer.

The pairwise DTW used for graph construction adds a per-window cost of $\binom{N}{2}$ comparisons on $\delta$-length snapshots, bounded by 1485 calls at our largest variable count $N\!=\!55$ with $\delta\!=\!10$, and the resulting graphs stay sparse at 1.14--3.33\% density (Tab.~\ref{tab:dataset_and_graph_construction}). For larger $N$, Sakoe-Chiba banding~\cite{sakoe1978dynamic} reduces each DTW call from $\mathcal{O}(\delta^2)$ to $\mathcal{O}(\delta w)$ with $w \ll \delta$, keeping the construction cost linear in the band width and well within the per-epoch budget.

\begin{figure*}[tbh]
  \centering  \includegraphics[width=\textwidth] {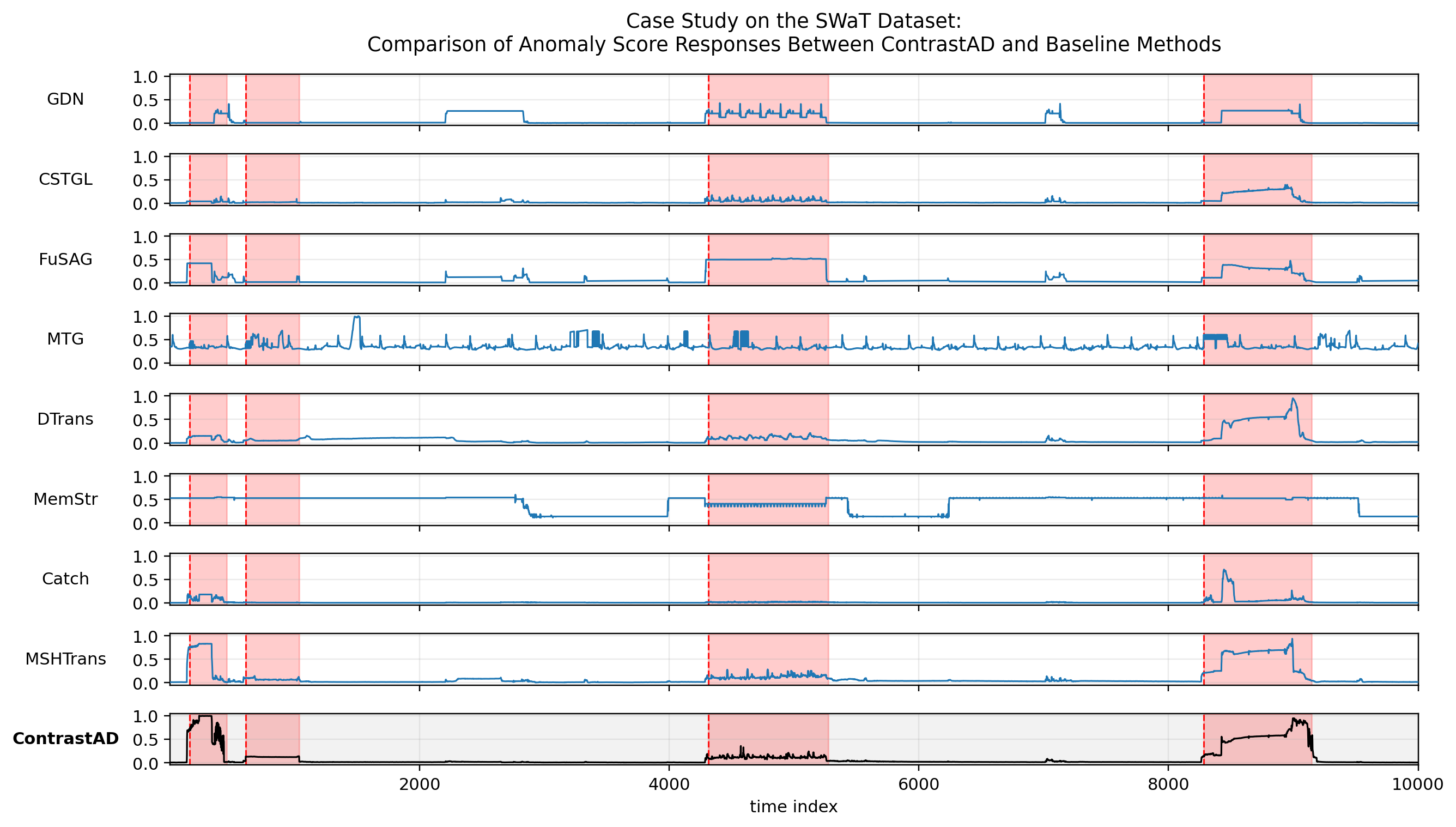} 
      \caption{Case study on the SWaT dataset comparing anomaly score responses of ContrastAD and baseline methods. Anomalies correspond to cyber-physical attacks annotated in the SWaT test set. The dashed red vertical line indicates the anomaly onset. All anomaly scores are independently min-max normalized to the range $[0,1]$ for visualization. The x-axis shows a sequence of 10{,}000 timestamp indices selected from the full test dataset.
    }
    \Description{Line plot showing anomaly score curves over 10,000 timestamps from the SWaT test set for ContrastAD and eight baseline methods. Each method's score is min-max normalized to the [0,1] range. The dashed red vertical line marks the anomaly onset. ContrastAD shows a sharp peak at anomaly onset and quickly returns to low values, while baselines show delayed, smoothed, or fluctuating responses.}

  \label{fig:case-study-baseline}
\end{figure*}

\subsubsection{RQ5. Case Study on SWaT}
Quantitative metrics aggregate performance across an entire test set but do not reveal whether a model reacts promptly at anomaly onset or maintains stable baseline scores during normal operation. Fig.~\ref{fig:case-study-baseline} shows anomaly score curves from ContrastAD and all eight baselines over a 10{,}000-step SWaT interval containing three annotated cyber-physical attack periods.

GDN and CSTGL produce near-zero scores throughout, missing the first two attacks and responding only weakly to the third. FuSAG fires a spurious peak around index 1500 in a normal window, a false alarm from sparse latent graph estimation. MTG stays near 0.5 across the interval with no separation between normal and anomalous windows, a sign of unstable density estimation. DTrans rises only in the third attack and largely misses the first two. MemStr holds a step-like elevated score throughout, indistinguishable from its normal baseline and unreliable for threshold-based detection. Catch responds clearly only to the third attack, as its frequency-patch reconstruction error rises only for sustained, magnitude-dominant anomalies. MSHTrans detects the first onset quickly but recovers slowly, staying elevated well after each attack ends.

ContrastAD is the only method that spikes sharply at all three onset markers, recovers to a low baseline between attacks, and stays noise-free during normal operation. The SWaT attacks here override actuator setpoints while the paired sensors still report a steady physical state, breaking the normal sensor-actuator coupling and producing a structural rather than a magnitude anomaly. DGCL contrasts the most divergent graph snapshot against a stable anchor at every window, so the score tracks this relational deviation directly, rising at onset and falling once the inter-variable structure returns to normal. The reconstruction-based baselines Catch and MemStr score by raw-signal error and respond only to magnitude shifts, consistent with Catch missing the first two short attacks and MemStr's permanently elevated baseline in Fig.~\ref{fig:case-study-baseline}.

\section{Conclusion}

We present ContrastAD, an unsupervised anomaly detection framework for multivariate time series that explicitly models evolving inter-variable dependencies. By integrating multi-perspective embeddings, frequency-aware attention fusion, and dynamic graph contrastive learning, ContrastAD effectively captures both temporal dynamics and structural evolution in non-stationary systems.

A key contribution of this work is the DGCL, which regularizes evolving graph structures through contrastive learning. Instead of enforcing strict invariance across snapshots, DGCL aligns stable relational patterns while accommodating structural drifts inherent in non-stationary systems. This approach ensures that the learned embeddings remain sensitive to the relational changes necessary for anomaly detection. Experimental results across five real-world benchmarks demonstrate that ContrastAD achieves the highest mean F1 on all five datasets, with statistically significant F1 and AUC improvements on SWaT and PSM against eight competitive baselines, while ablation results further indicate that contrastive learning is most effective when used as a soft regularizer, highlighting that rigid invariance for a static graph is suboptimal in dynamic environments.

In future work, we plan to explore adaptive contrastive weighting strategies
that respond to structural stability over time, compare alternative
time-series-to-graph constructions such as correlation-based or learnable
adjacency against the current DTW-based snapshot graph, investigate more
expressive snapshot-divergence measures such as spectral distances beyond the
current degree-dis\-tri\-bu\-tion KL, and apply the proposed framework to other
domains with dynamic relational patterns, such as financial systems and
network security.

\begin{acks}
This work was supported by UK Research and Innovation (UKRI) Engineering and Physical Sciences Research Council (EPSRC) [grant number EP/Y028392/1]: AI for Collective Intelligence (AI4CI).
\end{acks}

\ifnum\ARXIV=1 
\else
    \clearpage
\fi


\bibliographystyle{ACM-Reference-Format}
\bibliography{cikm26}

\end{document}